\RequirePackage{fix-cm}
\documentclass[10pt]{article}
\PassOptionsToPackage{dvipsnames,table}{xcolor}
\usepackage{yalenlp}

\usepackage{amsmath,amsfonts,bm}

\def\eqref#1{equation~\ref{#1}}

\def\1{\bm{1}}

\DeclareMathAlphabet{\mathsfit}{\encodingdefault}{\sfdefault}{m}{sl}
\SetMathAlphabet{\mathsfit}{bold}{\encodingdefault}{\sfdefault}{bx}{n}

\newcommand{\E}{\mathbb{E}}

\newcommand{\KL}{D_{\mathrm{KL}}}

\usepackage{url}
\usepackage{xspace}
\usepackage{algorithm}
\usepackage[noend]{algpseudocode}
\usepackage{amsmath,amssymb,mathtools,amsthm}
\usepackage{booktabs,multirow,array,tabularx,makecell}
\usepackage{graphicx}
\usepackage{xcolor}
\usepackage{microtype}
\usepackage{enumitem}
\usepackage{tikz}
\usetikzlibrary{arrows.meta,positioning,fit,calc,backgrounds}
\usepackage{pgfplots}
\pgfplotsset{compat=1.18}
\usepackage[nameinlink,capitalize,noabbrev]{cleveref}

\definecolor{EvidenceBlue}{HTML}{2F6B9A}
\definecolor{BypassOrange}{HTML}{D97925}
\definecolor{RepairGreen}{HTML}{2C8C69}
\definecolor{CounterRed}{HTML}{B94B55}
\definecolor{SoftBlue}{HTML}{EAF2F8}
\definecolor{SoftOrange}{HTML}{FCEFE4}
\definecolor{SoftGreen}{HTML}{EAF6F1}
\definecolor{SoftGray}{HTML}{F3F4F5}
\definecolor{TextGray}{HTML}{4C5661}

\newtheorem{theorem}{Theorem}
\newtheorem{proposition}[theorem]{Proposition}

\newtheorem{corollary}[theorem]{Corollary}
\theoremstyle{definition}

\newcommand{\method}{\textsc{EP-OPD}\xspace}
\newcommand{\qasper}{\textsc{Qasper}\xspace}
\newcommand{\opd}{\textsc{OPD}\xspace}
\newcommand{\base}{\textsc{Base}\xspace}
\newcommand{\cotsft}{\textsc{CoT-SFT}\xspace}
\newcommand{\anssft}{\textsc{Ans-SFT}\xspace}
\newcommand{\Dorig}{D^{\mathrm{orig}}}
\newcommand{\Dpara}{D^{\mathrm{para}}}
\newcommand{\Dirr}{D^{\mathrm{irr}}}
\newcommand{\Ddrop}{D^{\mathrm{drop}}}
\newcommand{\Dcf}{D^{\mathrm{cf}}}
\newcommand{\Acc}{\mathrm{Acc}}
\newcommand{\CFAcc}{\mathrm{CFAcc}}
\newcommand{\Bypass}{\mathrm{BypassRate}}

\newcommand{\score}{s_{\theta}}
\newcommand{\student}{\pi_{\theta}}
\newcommand{\teacher}{\pi_{T}}
\newcommand{\ind}{\mathbf{1}}
\newcommand{\loss}{\mathcal{L}}
\newcommand{\dataset}[1]{\textsc{#1}}
\newcommand{\best}[1]{\textbf{#1}}

\newcolumntype{Y}{>{\centering\arraybackslash}X}
\newcolumntype{L}[1]{>{\raggedright\arraybackslash}p{#1}}
\newcolumntype{C}[1]{>{\centering\arraybackslash}p{#1}}
\newcolumntype{R}[1]{>{\raggedleft\arraybackslash}p{#1}}

\setlist[itemize]{leftmargin=*,topsep=2pt,itemsep=1pt,parsep=0pt}
\setlist[enumerate]{leftmargin=*,topsep=2pt,itemsep=1pt,parsep=0pt}

\newcommand{\BaseOrig}{49.0}
\newcommand{\BasePara}{48.7}
\newcommand{\BaseCF}{44.5}
\newcommand{\BaseBypass}{20.1}
\newcommand{\BaseDrop}{22.0}
\newcommand{\BaseCSP}{69.8}

\newcommand{\AnsOrig}{50.2}
\newcommand{\AnsPara}{49.9}
\newcommand{\AnsCF}{43.8}
\newcommand{\AnsBypass}{23.7}
\newcommand{\AnsDrop}{19.2}
\newcommand{\AnsCSP}{67.5}

\newcommand{\CotOrig}{51.4}
\newcommand{\CotPara}{51.0}
\newcommand{\CotCF}{42.3}
\newcommand{\CotBypass}{29.6}
\newcommand{\CotDrop}{16.0}
\newcommand{\CotCSP}{64.2}

\newcommand{\OpdOrig}{52.4}
\newcommand{\OpdPara}{51.9}
\newcommand{\OpdCF}{39.2}
\newcommand{\OpdBypass}{35.8}
\newcommand{\OpdDrop}{14.4}
\newcommand{\OpdCSP}{57.0}

\newcommand{\EpOrig}{51.9}
\newcommand{\EpPara}{51.6}
\newcommand{\EpCF}{42.7}
\newcommand{\EpBypass}{25.9}
\newcommand{\EpDrop}{19.1}
\newcommand{\EpCSP}{65.4}

\newcommand{\MainOrigGain}{3.4}
\newcommand{\MainCFLoss}{-5.3}
\newcommand{\MainInteraction}{-8.7}

\newcommand{\MainBypassGain}{15.7}
\newcommand{\EpOrigCost}{-0.5}
\newcommand{\EpCFGain}{3.5}

\newcommand{\EpCSPGain}{8.4}
\newcommand{\EpGainRetained}{85}
\newcommand{\EpCFRecovered}{66}

\newcommand{\JudgeErrBase}{30.2}
\newcommand{\JudgeErrOpd}{43.0}
\newcommand{\JudgeErrDelta}{12.8}
\newcommand{\ExpertErrBase}{28.5}
\newcommand{\ExpertErrOpd}{38.3}
\newcommand{\ExpertErrDelta}{9.8}

\newcommand{\EmaSlowOrig}{54.8}
\newcommand{\EmaSlowCF}{41.5}
\newcommand{\EmaSlowBypass}{33.2}
\newcommand{\EmaSlowCSP}{59.1}

\newcommand{\EmaFastOrig}{51.6}
\newcommand{\EmaFastCF}{39.7}
\newcommand{\EmaFastBypass}{35.4}
\newcommand{\EmaFastCSP}{57.5}

\newcommand{\WeakPIOrig}{57.3}
\newcommand{\WeakPICF}{47.8}
\newcommand{\WeakPIBypass}{22.8}
\newcommand{\WeakPICSP}{68.2}

\newcommand{\StrongPIOrig}{60.6}
\newcommand{\StrongPICF}{52.6}
\newcommand{\StrongPIBypass}{14.8}
\newcommand{\StrongPICSP}{76.0}

\newcommand{\ForwardOrig}{50.7}
\newcommand{\ForwardCF}{43.1}
\newcommand{\ForwardBypass}{28.8}
\newcommand{\ForwardCSP}{63.2}

\newcommand{\SoftForwardOrig}{48.9}
\newcommand{\SoftForwardCF}{45.7}
\newcommand{\SoftForwardBypass}{24.5}
\newcommand{\SoftForwardCSP}{66.4}

\newcommand{\EntropyFiftyOrig}{47.2}
\newcommand{\EntropyFiftyCF}{39.8}
\newcommand{\EntropyFiftyBypass}{34.1}
\newcommand{\EntropyFiftyCSP}{58.1}

\newcommand{\EntropyTwentyFiveOrig}{43.8}
\newcommand{\EntropyTwentyFiveCF}{36.9}
\newcommand{\EntropyTwentyFiveBypass}{38.2}
\newcommand{\EntropyTwentyFiveCSP}{54.3}

\title{Better Answers, Weaker Evidence Control: Evidence Bypass in On-Policy Distillation}
\author{%
  \textbf{Ruoran Xu}\qquad\textbf{Yilun Zhao}\qquad\textbf{Ziyu Chen}\qquad\textbf{Xiaoqiang Kang}\qquad\textbf{Arman Cohan}\\[7pt]
  Yale NLP Lab
}
\date{}
\hypersetup{pdftitle={Better Answers, Weaker Evidence Control: Evidence Bypass in On-Policy Distillation},
            pdfauthor={Ruoran Xu, Yilun Zhao, Ziyu Chen, Xiaoqiang Kang, Arman Cohan}}

\begin{document}
\maketitle
\thispagestyle{yalefirst}

\begin{abstract}
On-policy distillation (OPD) can improve answer accuracy while weakening responsiveness to supporting evidence. We investigate this separation in scientific-document question answering through matched document interventions that change the supported answer while preserving the question and unrelated context. We identify evidence bypass: a student retains an originally correct answer after the document changes to support a different one. On QASPER with a Qwen3-4B student, standard OPD raises original-document accuracy by 3.4 percentage points, lowers counterfactual accuracy by 5.3 points, and increases bypass by 15.7 points; the same directional pattern appears on PeerQA. Controlled perturbations localize the shift to answer-critical evidence, and bypass grows with the student's pre-distillation closed-book preference for the original answer. A factorized study of teacher targets, privileged information, divergence, and supervision density finds that explicitly supplying the teacher with annotated answer-critical spans yields the largest joint improvement in evidence response and claim support. These findings motivate Evidence-Paired OPD (EP-OPD), which combines on-policy supervision on matched documents supporting different answers to the same question with an answer-level margin that favors each document's supported answer over its paired alternative. Across both datasets and two student sizes, EP-OPD recovers 54--66\% of standard OPD's counterfactual-accuracy loss while preserving 80--85\% of its original-document accuracy gain. Our results establish evidence responsiveness as a distinct training and evaluation target alongside answer accuracy. Our code is publicly available at: https://github.com/RuoranXu/EP-OPD.

\end{abstract}
\begin{figure}[!h]
    \centering
    \includegraphics[width=0.95\linewidth]{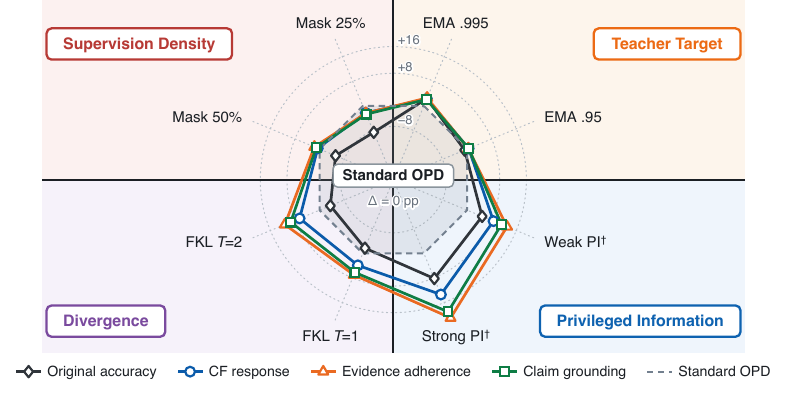}
    \vspace{-7pt}
    \caption{\textbf{A coordinate view of how ordinary OPD choices affect document use.}
    We vary supervision density, teacher target, divergence, and teacher privileged information one factor at a time around standard OPD (dashed contour). Each contour plots one metric's change (pp): original accuracy and the counterfactual response, evidence adherence, and claim grounding of the three panels of \cref{fig:opd-frontier}; outward is better. Teacher and density changes move the three evidence metrics almost identically, so those contours overlap. PI ($\dagger$): teacher privileged information.}
    \label{fig:design-compass}
    \vspace{-5pt}
\end{figure}

\section{Introduction}
\label{sec:intro}
Research on reasoning faithfulness distinguishes answer correctness and explanation plausibility from whether a model's stated reasons reflect the factors driving its decisions~\citep{ye2022unreliable,turpin2023unfaithful,chen2025reasoning}. Correct answers and coherent chains of thought (CoT) can coexist with a mismatch between stated justifications and actual decision influences. Interventions on reasoning traces and task inputs probe this relationship by measuring how predictions respond to changes in their stated basis~\citep{lanham2023measuring,atanasova2023faithfulness}.
Reasoning distillation brings this issue into training. Students learn from teacher-generated rationales and CoT traces~\citep{hsieh2023distilling,ho2023reasoningteachers,magister2023teaching}, raising a further question: does transferring reasoning behavior also transfer dependence on supporting evidence?

On-policy distillation (OPD) provides a setting for studying this question: the student generates its own trajectories, and the teacher supervises the prefixes it visits~\citep{agarwal2024gkd,gu2024minillm}. In document-grounded tasks, these trajectories bring the student's existing answer preferences into contact with evidence-conditioned teacher supervision. 
We investigate this question in scientific-document QA. Document evidence and parametric answer preferences often support the same answer on ordinary examples. We examine their relationship through controlled document variation: we change one answer-critical fact and update dependent mentions, producing a matched document that supports a different answer. Fresh responses to both versions measure how distillation changes the student's responsiveness to supporting evidence.

We identify evidence bypass: a model retains an originally correct answer after the document changes to support a different one. On \qasper with a Qwen3-4B student, standard OPD increases original-document accuracy from \BaseOrig\% to 52.4\%, while counterfactual accuracy falls from 44.5\% to 39.2\% and bypass rises from 20.1\% to 35.8\%. The same directional pattern appears on PeerQA. These results reveal a separation between improvements in answer accuracy and changes in evidence dependence during distillation.

Further analysis connects this pattern to answer-critical information and pre-existing answer preferences. Paraphrases and irrelevant insertions preserve most of OPD's original-document gain, while evidence deletion reduces correct abstention.
To examine how the training signal shapes evidence dependence, we vary four OPD design dimensions: teacher target, teacher privileged information, divergence loss, and supervision density. Among these interventions, providing the teacher with the annotated answer-critical span yields the largest gains in counterfactual accuracy and claim support, together with the largest reduction in bypass.

These findings motivate Evidence-Paired OPD (EP-OPD), which makes document-dependent answer variation explicit during training. Each question is paired with two documents supporting different answers. Both receive on-policy teacher supervision, and a paired answer-level objective encourages the student to favor each document's supported answer over its paired alternative. Across \qasper and PeerQA and two student sizes, EP-OPD recovers 54--66\% of standard OPD's counterfactual-accuracy loss while preserving 80--85\% of its original-document accuracy gain. These results establish evidence dependence as a training and evaluation target alongside answer accuracy.

Our contributions are threefold:
\begin{itemize}
    \item We identify evidence bypass in document-grounded OPD through matched evidence interventions and characterize its relationship to the student's pre-distillation answer preferences.
    \item We organize OPD design along teacher target, teacher privileged information, divergence loss, and supervision density, showing how each shapes accuracy, evidence response, and claim support.
    \item We introduce Evidence-Paired OPD, which couples on-policy supervision with document-dependent answer preferences to strengthen evidence response while preserving most of distillation's accuracy gains.
\end{itemize}
\section{Related work}
\label{sec:related}
\subsection{Chain-of-thought faithfulness and evidence interventions}
A correct answer accompanied by a plausible chain of thought (CoT) does not establish that the stated reasoning faithfully reflects the model's decision process. \citet{ye2022unreliable} show that generated explanations can lack input grounding. \citet{turpin2023unfaithful} demonstrate that CoT traces can omit biasing features influencing answers, while \citet{chen2025reasoning} document analogous omissions of influential hints in reasoning models. Interventions on reasoning traces and task inputs probe whether predictions depend on their stated basis~\citep{lanham2023measuring,atanasova2023faithfulness}, complementing evaluations of rationale sufficiency, comprehensiveness, and explanation--label association~\citep{deyoung2020eraser,wiegreffe2021association}.
Counterfactual data and contrast sets probe decision rules through targeted input changes~\citep{kaushik2020counterfactual,gardner2020contrastsets}, while attribution research assesses whether claims are supported by source material~\citep{rashkin2023attribution,gao2023citations}. We apply these perspectives to scientific-document QA using answer-changing document pairs, evidence deletion, and claim-support evaluation. These diagnostics measure behavioral evidence responsiveness rather than directly verify internal reasoning.
\subsection{On-policy and reasoning distillation}
Reasoning distillation transfers teacher behavior through generated rationales and CoT traces~\citep{hsieh2023distilling,ho2023reasoningteachers,magister2023teaching}. On-policy distillation (OPD) supervises student-generated trajectories: GKD applies teacher distributions at student-visited states~\citep{agarwal2024gkd}, MiniLLM optimizes reverse KL using on-policy samples~\citep{gu2024minillm}, and \citet{li2026rethinking} analyze progressive teacher--student alignment along these trajectories.
CoT faithfulness raises a further question: does imitating teacher reasoning also transfer dependence on supporting evidence? Agreement on observed answers and traces alone need not establish such dependence. SCOTT addresses a related rationale-level issue through counterfactual rationale--answer pairs~\citep{wang2023scott}. We instead examine document-level evidence dependence under OPD, varying teacher target, teacher privileged information, divergence loss, and supervision density. Evidence-Paired OPD (EP-OPD) trains on matched documents supporting different answers to the same question, coupling answer preferences to source evidence rather than varying supplied rationales.
\subsection{Context faithfulness and parametric knowledge}
Knowledge-conflict research examines how models balance supplied context against parametric knowledge. Entity substitution and counterfactual QA probe reliance on these sources~\citep{longpre2021entity,neeman2023disentqa}. FaithEval evaluates context-following under controlled evidence conditions~\citep{ming2025faitheval}, and Context-DPO trains preferences for context-faithful answers~\citep{bi2025contextdpo}. Scientific-reasoning research further shows that stronger reasoning can make existing parametric knowledge more accessible~\citep{li2026demystifying}. We connect these observations to distillation by relating the base student's closed-book answer preference to post-OPD responses under paired document interventions, without interpreting this association as direct identification of an internal reasoning mechanism.
\section{Experimental Framework}
\label{sec:framework}

We separate a model's accuracy on the supplied document from its response when answer-critical evidence changes, and compare how training alters each. This section defines the matched document interventions, the metrics and statistical comparisons built on them, and the models and OPD design variations we compare.

\subsection{Datasets and Document Interventions}
\label{sec:data-interventions}

We evaluate on two scientific document QA datasets: \qasper comprises 5,049 questions over 1,585 NLP papers~\citep{dasigi2021qasper}, and PeerQA comprises 579 questions drawn from reviews of 208 papers~\citep{baumgartner2025peerqa}. Both provide answers and supporting evidence, with distinct question-collection procedures and answer formats. Splits are disjoint by source paper. The primary evaluation includes answerable questions whose answers can be represented as an entity, value, polarity, comparison, or bounded set; unanswerable examples contribute to abstention evaluation. We use full papers when they fit the context window. Otherwise, a fixed document pack combines annotated evidence with retrieval-ranked distractor paragraphs and is shared across methods.

We measure document response by changing the evidence that determines an answer while keeping the question fixed. For each eligible example, we edit one answer-critical span, changing an entity, value, comparison, polarity, or reported direction. We update dependent mentions consistently and preserve the question, section order, and unrelated context:

\begin{equation}
  (q,\Dorig,a^{\mathrm{orig}})
  \longrightarrow
  (q,\Dcf,a^{\mathrm{cf}}),
  \qquad a^{\mathrm{orig}}\neq a^{\mathrm{cf}}.
  \label{eq:intervention}
\end{equation}
We generate a fresh response to each document version and validate each held-out pair so that the edited document uniquely supports its new answer. We also construct a meaning-preserving paraphrase $\Dpara$, an evidence-deleted version $\Ddrop$, and a length-matched irrelevant insertion $\Dirr$. The answer-changing edit preserves answer type and scientific syntax and updates dependent mentions. Validators check label uniqueness, residual support for the original answer, and edit locality; held-out pairs also receive manual review before model evaluation. These versions measure response to semantic preservation, evidence availability, distraction, and answer-changing evidence (\cref{app:construction}).

\subsection{Evidence-Response Metrics and Evaluation}
\label{sec:metrics}

Three primary quantities summarize answer behavior. \emph{Original accuracy} measures correctness on $\Dorig$; \emph{counterfactual accuracy} (CFAcc) measures correctness on $\Dcf$. \emph{Bypass} is computed over the examples each model answers correctly on $\Dorig$: the fraction on which the same model retains the original answer on $\Dcf$. Because this conditioning set is model-specific, bypass differences compare conditional rates rather than changes on a fixed question set. The three measures capture performance on the observed document, adaptation to changed evidence, and persistence of the original answer. CFAcc and bypass are complementary: avoiding the original answer does not by itself mean producing the answer the edited document supports.

We also measure correct abstention after evidence deletion. We score original accuracy, counterfactual accuracy, and bypass with deterministic, item-specific answer canonicalizers. We split responses into independently evaluable claims and assess source support. A blinded evaluator receives the question, one randomized claim, its cited spans, and minimal neighboring context, with method and outcome metadata hidden. Labels are supported, contradicted, absent from the source, invalid composition, or irrelevant. Claim support precision (CSP) is the fraction of answer-relevant claims labeled supported. Experts evaluate a disjoint subset using the same units and definitions. Segmentation rules, prompts, and evaluation controls appear in \cref{app:evaluation}. \Cref{app:formal-comparison} gives the metric definitions, paired interaction estimand, and regression form.

Each matched question pair is the comparison unit, and source papers define resampling clusters. We aggregate claim labels by question and report scores averaged across seeds. The primary comparison estimates how the training-induced change in correctness differs between counterfactual and original documents using $10{,}000$ paired, paper-clustered bootstrap replicates. We apply Holm correction to secondary outcomes within each result family.

\subsection{Models and Factorized OPD Design}
\label{sec:opd-design}

The phenomenon analysis uses Qwen3-4B base students and a frozen Qwen3-30B-A3B-Instruct-2507 teacher~\citep{yang2025qwen3}. Training comparisons span the pretrained \base checkpoint, answer-only supervision (\anssft), imitation of teacher reasoning traces whose final answers pass the task verifier (answer-filtered \cotsft), and standard reverse-KL \opd. Principal runs share the question pool, optimizer-step count, retrieved context, and supervised output-token budget. We update all student parameters and run three seeds per principal configuration. Hyperparameters and compute matching appear in \cref{app:training}.

We factorize standard OPD along four dimensions, teacher target, teacher privileged information, divergence, and supervision density, and vary one dimension at a time under an otherwise shared training setup. Teacher target: fixed reference or EMA with $\mu\in\{0.995,0.95\}$. Teacher privileged information: gold evidence IDs or the answer-critical span. Divergence: reverse KL or forward KL with $T\in\{1,2\}$. Supervised positions: all prefixes or the top $50\%$ or $25\%$ by pre-update student entropy. Questions, rollout settings, optimizer schedules, and student inputs remain fixed within each comparison. Exact objectives appear in \cref{app:factor-controls}.

Teacher privileged information is supplied only to the teacher; student rollouts and evaluation inputs never contain it. Entropy-mask controls hold rollouts and optimizer steps fixed, while the number of teacher-scored tokens intentionally varies with supervision density (\cref{app:factor-controls}). Results of the factorized study are reported in \cref{sec:opd-design-results} and \cref{tab:opd-factor-controls}.

\section{Evidence Bypass under On-Policy Distillation}
\label{sec:bypass}

We examine how standard OPD changes evidence response in Qwen3-4B students, establishing the relationship between original-document accuracy and original-answer persistence under counterfactual evidence. We then characterize this change through document interventions and pre-distillation answer preferences, before examining how evidence response varies across the OPD design choices defined in \cref{sec:opd-design}.

\subsection{Accuracy Gains Accompany Stronger Original-Answer Persistence}

\begin{table}[!t]
\caption{\qasper with Qwen3-4B-Base (\%). $\uparrow$/$\downarrow$: higher/lower is better. Among the baseline training conditions, standard OPD attains the highest original and paraphrase accuracy while every evidence-sensitive outcome moves away from the base model. \method (last row; \cref{sec:epopd-evaluation}) partially recovers these outcomes at a 0.5\,pp original-accuracy cost.}
\label{tab:qasper-main}
\centering
\small
\setlength{\tabcolsep}{4.4pt}
\begin{tabular}{lrrrrrr}
\toprule
\textbf{Method} & \textbf{Orig. $\uparrow$} & \textbf{Para. $\uparrow$} & \textbf{CFAcc $\uparrow$} & \textbf{Bypass $\downarrow$} & \textbf{Drop abst. $\uparrow$} & \textbf{CSP $\uparrow$} \\
\midrule
\base & \BaseOrig & \BasePara & \best{\BaseCF} & \best{\BaseBypass} & \best{\BaseDrop} & \best{\BaseCSP} \\
\anssft & \AnsOrig & \AnsPara & \AnsCF & \AnsBypass & \AnsDrop & \AnsCSP \\
Answer-filtered \cotsft & \CotOrig & \CotPara & \CotCF & \CotBypass & \CotDrop & \CotCSP \\
Standard \opd & \best{\OpdOrig} & \best{\OpdPara} & \OpdCF & \OpdBypass & \OpdDrop & \OpdCSP \\
\midrule
\method (ours) & \EpOrig & \EpPara & \EpCF & \EpBypass & \EpDrop & \EpCSP \\
\bottomrule
\end{tabular}
\end{table}

On \qasper, standard OPD increases original-document accuracy from \BaseOrig\% to 52.4\%. On the matched answer-changing documents, counterfactual accuracy falls from 44.5\% to 39.2\%, and conditional bypass rises from 20.1\% to 35.8\%. The paired interaction is $-8.7$ percentage points (95\% CI, $[-12.4,-5.1]$), capturing the separation between the training-induced changes on original and counterfactual documents. OPD therefore improves accuracy on the original documents while increasing persistence of the original answer when the supplied evidence supports a different one.

The supervised baselines place this pattern in the context of different training signals. Answer-only SFT produces modest increases in original accuracy and bypass; answer-filtered CoT-SFT produces larger increases in both; and standard OPD exhibits the largest original-document gain and bypass increase among these baseline training conditions (first four rows of \cref{tab:qasper-main}). The same direction appears on PeerQA with the 4B student: standard OPD raises original accuracy by 2.1 percentage points and lowers counterfactual accuracy by 2.6 points, yielding a paired interaction of $-4.7$ points. Across both datasets, the matched evaluations reveal a consistent shift toward retaining the original answer after an answer-changing edit.

\subsection{Document Interventions Localize the Shift to Answer-Critical Evidence}

The document controls connect this shift to the information needed to answer the question. On \qasper, accuracy on meaning-preserving paraphrases rises from 48.7\% to 51.9\%, closely tracking the original-document improvement. Irrelevant insertions likewise preserve most of the original-document gain. Removing the necessary evidence produces a larger change: correct abstention falls from 22.0\% to 14.4\%. The accuracy gain thus carries over to edits that preserve the answer, whereas abstention under evidence deletion and accuracy under answer-changing edits both decline.

Claim-level evaluation extends the analysis to the support for individual answer-relevant statements. Automatic evidence error rises from 30.2\% to 43.0\%, and error on the blinded expert subset rises from 28.5\% to 38.3\%. These changes accompany a 12.8-point reduction in claim support precision. \Cref{tab:grounding-controls} brings together the document-operator and claim-level outcomes, linking the change in final-answer behavior to weaker support for the claims used in the response.

\subsection{Bypass Increases with Pre-Distillation Answer Preference}


The additional bypass induced by OPD grows monotonically across the resulting prior-strength quartiles, from approximately 5 percentage points in the weakest-prior quartile to 20 points in the strongest. Across the two datasets, the three-way interaction between OPD, counterfactual evaluation, and high prior strength lies between $-7$ and $-10$ points. The strongest change in evidence response thus occurs where the base student already favors the original answer.

The document interventions and prior stratification provide complementary views of evidence bypass: the former locate the shift in answer-critical evidence, and the latter relate its magnitude to an existing answer preference. We next examine how evidence response changes when individual OPD design choices are varied.

\subsection{Evidence Response under OPD Design Variations}
\label{sec:opd-design-results}

The preceding analysis concerns standard OPD. We now ask whether the separation between original accuracy and evidence response persists, and how it changes, when a single training choice is varied. Using the factorized design in \cref{sec:opd-design}, we vary one dimension at a time on \qasper with Qwen3-4B. \Cref{tab:opd-factor-controls} reports the results, and \cref{fig:design-compass} gives an overview.

\begin{table}[!t]
\caption{Factorized OPD interventions on \qasper with Qwen3-4B (\%). Each intervention changes one design dimension relative to standard OPD. $^{\dagger}$Gold evidence information is provided only to the teacher.}
\label{tab:opd-factor-controls}
\centering
\scriptsize
\setlength{\tabcolsep}{5.0pt}
\begin{tabular}{llrrrr}
\toprule
\textbf{Axis} & \textbf{Single change} & \textbf{Orig. $\uparrow$} & \textbf{CFAcc $\uparrow$} & \textbf{Bypass $\downarrow$} & \textbf{CSP $\uparrow$} \\
\midrule
Reference & Standard OPD & \OpdOrig & \OpdCF & \OpdBypass & \OpdCSP \\
\midrule
Teacher & EMA, $\mu=0.995$ & \EmaSlowOrig & \EmaSlowCF & \EmaSlowBypass & \EmaSlowCSP \\
Teacher & EMA, $\mu=0.95$ & \EmaFastOrig & \EmaFastCF & \EmaFastBypass & \EmaFastCSP \\
Priv.\ info. & Weak PI: evidence IDs$^{\dagger}$ & \WeakPIOrig & \WeakPICF & \WeakPIBypass & \WeakPICSP \\
Priv.\ info. & Strong PI: critical span$^{\dagger}$ & \StrongPIOrig & \StrongPICF & \StrongPIBypass & \StrongPICSP \\
Divergence & Forward KL, $T=1$ & \ForwardOrig & \ForwardCF & \ForwardBypass & \ForwardCSP \\
Divergence & Forward KL, $T=2$ & \SoftForwardOrig & \SoftForwardCF & \SoftForwardBypass & \SoftForwardCSP \\
Density & Top-entropy $50\%$ & \EntropyFiftyOrig & \EntropyFiftyCF & \EntropyFiftyBypass & \EntropyFiftyCSP \\
Density & Top-entropy $25\%$ & \EntropyTwentyFiveOrig & \EntropyTwentyFiveCF & \EntropyTwentyFiveBypass & \EntropyTwentyFiveCSP \\
\bottomrule
\end{tabular}
\end{table}

Teacher-target changes remain close to standard OPD: a slow EMA target reaches \EmaSlowOrig\% original accuracy and \EmaSlowCF\% CFAcc, with \EmaSlowBypass\% bypass and \EmaSlowCSP\% CSP. A faster EMA target produces a smaller shift. Forward KL traces a different accuracy-response profile: moving from $T=1$ to $T=2$ raises CFAcc from \ForwardCF\% to \SoftForwardCF\% and reduces bypass from \ForwardBypass\% to \SoftForwardBypass\%, as original accuracy moves from \ForwardOrig\% to \SoftForwardOrig\%. Concentrating supervision on fewer high-entropy prefixes progressively lowers original accuracy.

Teacher privileged information produces the largest joint movement across the evidence-sensitive measures. Providing gold evidence IDs raises CFAcc to \WeakPICF\%, lowers bypass to \WeakPIBypass\%, and raises CSP to \WeakPICSP\%. Providing the exact answer-critical span moves these quantities to \StrongPICF\%, \StrongPIBypass\%, and \StrongPICSP\%, with original accuracy reaching \StrongPIOrig\%. Among the tested interventions, supplying answer-critical evidence to the teacher yields the largest joint improvement in CFAcc, bypass, and CSP.


\section{Evidence-Paired OPD}
\label{sec:method}
When each question is paired with only one answer-supporting document during training, evidence-responsive and document-insensitive strategies may behave identically on the observed examples (\cref{prop:nonident}). We therefore pair each question with documents supporting different answers, so that document-insensitive answer preferences cannot simultaneously satisfy both supervision targets.

\begin{figure}[!t]
\centering
\includegraphics[width=0.9\linewidth]{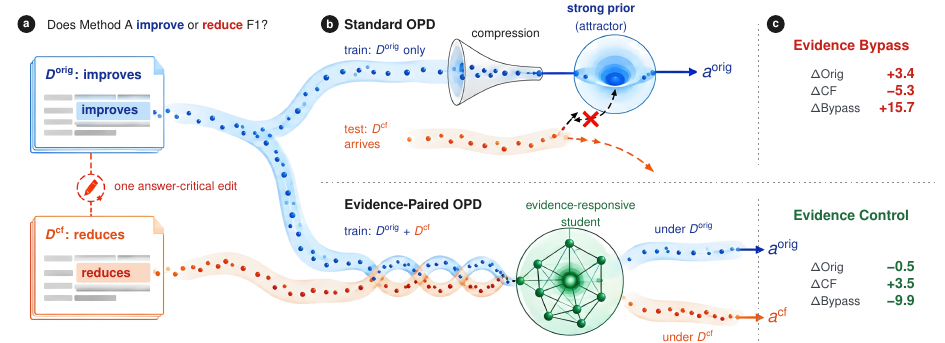}
\vspace{-10pt}
\caption{\textbf{Minimal evidence interventions identify bypass; paired evidence repairs it.}
One answer-critical edit flips the supported answer while the question and document structure stay fixed. Standard OPD sees only $D^{\mathrm{orig}}$ and may preserve $a^{\mathrm{orig}}$ under $D^{\mathrm{cf}}$; \method co-distills the pair and follows the edit. Values are Qwen--4B; deltas are relative to Base(top) and standard OPD(bottom).}
\label{fig:overview}
\end{figure}

Evidence-Paired OPD (\method) turns the matched-document intervention into a training signal. Each question appears with two validated documents that support distinct answers. The student receives on-policy teacher supervision under each document, together with an answer-level objective that couples its answer preferences to the evidence supplied in that document. \Cref{fig:overview} shows the document intervention and the associated training outcomes.

\subsection{Paired-document objective}

For a validated pair, let $x^0=(q,D^0)$ and $x^1=(q,D^1)$ support answers $a^0\neq a^1$. The student independently samples a response under each input. At every student-visited prefix, the frozen teacher conditions on the same question and document and supplies the next-token distribution used by the standard reverse-KL OPD objective.

Let $s_\theta(a\mid x)$ denote the length-normalized log probability of canonical answer $a$. We encourage each document to favor its supported answer over the answer associated with its partner:
\begin{equation}
\mathcal{L}_{\mathrm{pair}}
=
\sum_{b\in\{0,1\}}
\left[
m-s_\theta(a^b\mid x^b)
+s_\theta(a^{1-b}\mid x^b)
\right]_+,
\label{eq:pairloss}
\end{equation}
where $m>0$ is the answer margin and $[z]_+=\max\{z,0\}$. The full objective is
\[
\mathcal{L}_{\mathrm{EP\text{-}OPD}}
=
\frac{1}{2}\sum_{b\in\{0,1\}}
\mathcal{L}_{\mathrm{OPD}}(x^b;\theta)
+\lambda\mathcal{L}_{\mathrm{pair}}.
\]
The OPD term transfers teacher behavior along student-generated trajectories, and the paired term couples answer preferences to document variation. When the pair loss is zero, each supported answer exceeds its paired alternative in score by at least $m$ under its own document. The two canonical answers consequently exchange their ranking across the pair. Appendix~\ref{a3} formalizes this property, and \cref{app:epopd-details} gives the token-level loss and answer-score definitions.
\begin{algorithm}[H]
\caption{Evidence-Paired On-Policy Distillation}
\label{alg:epopd}
\small
\algrenewcommand{\algorithmicindent}{1em}
\begin{algorithmic}[1]
\Require Paired data $\mathcal{P}$, student $\student$,
         frozen teacher $\teacher$.
\Require Margin $m$, weight $\lambda$, batch size $B$,
         learning rate $\eta$.

\For{each training iteration}
    \State Sample $\mathcal{B} \subset \mathcal{P}$,
    $|\mathcal{B}|=B$; set $\widehat{\mathcal{L}} \gets 0$.

    \For{each $(q,D^0,a^0,D^1,a^1) \in \mathcal{B}$}
        \For{$b \in \{0,1\}$}
            \State $x^b \gets (q,D^b)$; sample
            $y^b \sim \student(\cdot \mid x^b)$.

            \State
            $p_t \gets \student(\cdot \mid x^b,y^b_{<t})$,
            $r_t \gets \teacher(\cdot \mid x^b,y^b_{<t})$
            for all $t$.

            \State
            $\ell_b \gets
            \frac{1}{|y^b|}
            \sum_{t=1}^{|y^b|}
            \sum_{v \in \mathcal{V}}
            p_t(v)\log\frac{p_t(v)}{r_t(v)}$.

            \For{$c \in \{0,1\}$}
                \State
                $s_{bc} \gets
                \frac{1}{|a^c|}
                \sum_{j=1}^{|a^c|}
                \log\student(a_j^c \mid x^b,a^c_{<j})$.
            \EndFor
        \EndFor

        \State
        $\mathcal{L}_{\mathrm{pair}} \gets
        [m-s_{00}+s_{01}]_{+}
        +[m-s_{11}+s_{10}]_{+}$.

        \State
        $\widehat{\mathcal{L}} \gets
        \widehat{\mathcal{L}}
        +\frac{1}{B}
        \bigl[
            \frac{\ell_0+\ell_1}{2}
            +\lambda\mathcal{L}_{\mathrm{pair}}
        \bigr]$.
    \EndFor

    \State Update $\theta$ using
    $\nabla_{\theta}\widehat{\mathcal{L}}$
    at learning rate $\eta$.
\EndFor

\Ensure Updated student parameters $\theta$.
\end{algorithmic}
\end{algorithm}

\subsection{Training procedure}

We construct and validate document pairs before optimization and place both members of each pair in the same mini-batch. Each member receives an independent student rollout and the corresponding teacher supervision. The paired answer scores are computed by teacher forcing over the short canonical answer strings. The margin and pair weight are selected on the \qasper development set and transferred unchanged to the second student size and PeerQA.

At inference, the trained student receives a single question--document input and generates its response. \Cref{alg:epopd} summarizes the optimization procedure. Appendix~\ref{app:training} specifies the hyperparameters and supervision-budget matching.

\section{Evaluating Evidence-Paired OPD}
\label{sec:epopd-evaluation}

We evaluate \method on held-out document pairs using the protocol in \cref{sec:framework}. The main comparison uses the Qwen3-4B student from the phenomenon analysis; the cross-dataset and scale study additionally includes Qwen3-1.7B. Details appear in Appendix~\ref{c3}.

\subsection{Answer accuracy and evidence response}

%
On \qasper with Qwen3-4B (\cref{tab:qasper-main}), \method achieves 51.9\% original-document accuracy, preserving \EpGainRetained\% of the gain obtained by standard OPD over the base student. The original-accuracy change relative to standard OPD is \EpOrigCost\ percentage points, and paraphrase accuracy is 51.6\%. Counterfactual accuracy increases by \EpCFGain\ points to 42.7\%, recovering \EpCFRecovered\% of standard OPD's counterfactual-accuracy loss. Conditional bypass falls from 35.8\% to 25.9\%, a reduction of 9.9 points.

The improvement extends to evidence availability and claim support. Correct abstention after evidence deletion rises by 4.7 points to 19.1\%, and claim support precision rises by \EpCSPGain\ points to 65.4\%. Paired training therefore improves adaptation to answer-changing evidence, abstention when necessary evidence is missing, and support for answer-relevant claims while preserving most of the original-document accuracy gain.

\begin{figure}[!t]
\centering
\includegraphics[width=\linewidth]{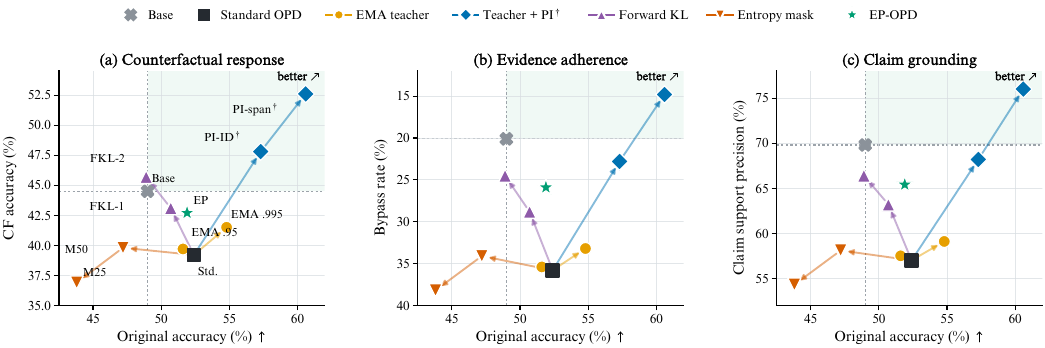}
\vspace{-4pt}
\caption{\textbf{One-axis OPD interventions trace different accuracy--evidence frontiers.}
Panels pair original accuracy with CFAcc, reversed bypass, or CSP (top-right is better). Arrows change one factor from standard OPD; dashed guides mark Base. PI ($\dagger$) is an oracle and \method a symmetric-context reference.}
\label{fig:opd-frontier}
\end{figure}

\Cref{fig:opd-frontier} places this response profile alongside the factorized OPD controls from \cref{sec:opd-design-results}. Teacher targets, divergence choices, and supervision density yield different combinations of original accuracy and evidence response. The teacher-side privileged-information controls measure the effect of supplying annotated evidence directly to the teacher. \method introduces variation through matched training documents, with teacher and student sharing the document input for each member. Across the three panels, paired training improves counterfactual response, evidence adherence, and claim grounding relative to standard OPD.

\subsection{Consistency across datasets and student sizes}

\Cref{tab:cross-setting} reports the standard-OPD change from the base student in each dataset--student setting and the fraction of its counterfactual-accuracy loss recovered by \method. On \qasper, paired training recovers 59\% of the counterfactual loss for the 1.7B student and 66\% for the 4B student. The corresponding recovery on PeerQA is 62\% and 54\%. Across the four settings, \method retains 80--85\% of the original-document accuracy gain while recovering 54--66\% of the counterfactual loss. The improvement in document response thus extends across both scientific QA datasets and both student sizes under the shared paired-training configuration.

\begin{table}[!t]
\caption{Cross-dataset and scale replication. Entries are percentage-point changes of standard OPD relative to the frozen base, except the final column, which is the fraction of OPD's CFAcc loss recovered by \method.}
\label{tab:cross-setting}
\centering
\begin{tabular}{llrrr}
\toprule
\textbf{Dataset} & \textbf{Student} & \textbf{$\Delta$Orig.} & \textbf{$\Delta$CFAcc} & \textbf{CF recovery} \\
\midrule
\qasper & 1.7B   & $+2.6$ & $-3.4$  & $59\%$ \\
\qasper & 4B & $+3.4$ & $-5.3$ &  $66\%$ \\
\dataset{PeerQA} & 1.7B & $+2.9$ & $-4.5$ & $62\%$ \\
\dataset{PeerQA} & 4B   & $+2.1$ & $-2.6$ & $54\%$ \\
\bottomrule
\end{tabular}
\end{table}

\subsection{Evidence response across prior-strength strata}

The frozen-base prior stratification connects the effect of pairing to the example-level pattern established in the phenomenon analysis. As shown in Appendix~\ref{f2}, \method lowers bypass in every quartile. In the weakest-prior quartile, bypass falls from 17\% under standard OPD to 14\%; in the strongest, it falls from 49\% to 35\%. Relative to the base student, these values correspond to additional bypass of 2 and 6 percentage points, compared with 5 and 20 points under standard OPD. Pairing thus improves evidence following throughout the prior distribution, with the largest absolute reduction in the strongest-prior quartile.

\paragraph{Component ablation.}
We compare three paired-document variants on \qasper with Qwen3-4B:
Augmented OPD, Paired ANS-SFT, and Augmented OPD with answer
cross-entropy. More details in Appendix~\ref{app:paired_ablation}.
Relative to Augmented OPD, EP-OPD improves CFAcc by 4.2 percentage
points and CSP by 3.1 points, with a 0.2-point decrease in original
accuracy. Replacing answer cross-entropy with the paired margin
improves CFAcc and CSP by 1.8 and 5.7 points, respectively, with
original accuracy changing from 52.5\% to 51.9\%.
Paired ANS-SFT attains the lowest bypass rate (25.3\%), while EP-OPD
achieves higher original accuracy, CFAcc, and CSP.

\section{Conclusion}
\label{sec:con}
We identify evidence bypass in document-grounded on-policy distillation: students improve their accuracy on original documents while increasingly retaining the original answer when the document supports a different one. Matched interventions on \qasper and PeerQA reveal that this shift tracks answer-critical evidence and grows with the student's pre-distillation preference for the original answer. We decompose standard OPD along four dimensions---teacher target, teacher privileged information, divergence loss, and supervision density---and find that teacher-side oracle access to answer-critical evidence yields the largest joint improvement in evidence-sensitive outcomes.

Building on these findings, Evidence-Paired OPD trains on matched documents that support different answers to the same question, explicitly coupling answer preferences to document evidence. It improves counterfactual accuracy and claim support, reduces bypass, and preserves most of standard OPD's original-document accuracy gain. 
Collectively, these results establish evidence response as a distinct dimension of distillation quality and motivate training objectives that strengthen both answer accuracy and responsiveness to answer-relevant evidence.

\subsection*{AI use statement}
Generative AI tools were used solely for limited assistance with language polishing, formatting, routine coding, and debugging. The research, analysis, interpretation of results, and preparation of the final content were carried out by the human authors. All results and citations were independently verified by the authors, who take full responsibility for the work.

\subsection*{Ethics statement}
Counterfactual documents are controlled research stimuli, not claims about the cited papers. They are clearly marked in released metadata and must not be indexed or redistributed as authentic scientific text. Human validation uses only document excerpts and model outputs; no sensitive personal data are collected. The method may improve evidence following, but it does not guarantee factuality when supplied evidence is itself false or adversarial, so counterfactual compliance should not be treated as a universal trustworthiness criterion.

\subsection*{Reproducibility statement}
The paper specifies the document interventions, metrics, statistical unit, resampling protocol, datasets, models, and baselines (\cref{sec:framework}), the training objective and algorithm (\cref{sec:method}), and the formal estimand (\cref{app:formal-comparison}). The appendix provides proofs, construction validators, evaluation prompts, hyperparameters, and additional result tables. The source package includes a summary result manifest and uses shared macros for the main reported outcomes. An anonymous implementation and generated intervention manifests will accompany the submission.



\bibliographystyle{unsrtnat}
\bibliography{references}

@inproceedings{agarwal2024gkd,
  title     = {On-Policy Distillation of Language Models: Learning from Self-Generated Mistakes},
  author    = {Agarwal, Rishabh and Vieillard, Nino and Zhou, Yongchao and Stanczyk, Piotr and Ramos Garea, Sabela and Geist, Matthieu and Bachem, Olivier},
  booktitle = {International Conference on Learning Representations},
  year      = {2024},
  url       = {https://openreview.net/forum?id=3zKtaqxLhW}
}

@inproceedings{baumgartner2025peerqa,
  title     = {{PeerQA}: A Scientific Question Answering Dataset from Peer Reviews},
  author    = {Baumgaertner, Tim and Briscoe, Ted and Gurevych, Iryna},
  booktitle = {Proceedings of the 2025 Conference of the Nations of the Americas Chapter of the Association for Computational Linguistics: Human Language Technologies},
  pages     = {508--544},
  year      = {2025},
  publisher = {Association for Computational Linguistics},
  doi       = {10.18653/v1/2025.naacl-long.22}
}

@inproceedings{bi2025contextdpo,
  title     = {{Context-DPO}: Aligning Language Models for Context-Faithfulness},
  author    = {Bi, Baolong and Huang, Shaohan and Wang, Yiwei and Yang, Tianchi and Zhang, Zihan and Huang, Haizhen and Mei, Lingrui and Fang, Junfeng and Li, Zehao and Wei, Furu and Deng, Weiwei and Sun, Feng and Zhang, Qi and Liu, Shenghua},
  booktitle = {Findings of the Association for Computational Linguistics: ACL 2025},
  pages     = {10280--10300},
  year      = {2025},
  publisher = {Association for Computational Linguistics},
  doi       = {10.18653/v1/2025.findings-acl.536}
}

@inproceedings{dasigi2021qasper,
  title     = {A Dataset of Information-Seeking Questions and Answers Anchored in Research Papers},
  author    = {Dasigi, Pradeep and Lo, Kyle and Beltagy, Iz and Cohan, Arman and Smith, Noah A. and Gardner, Matt},
  booktitle = {Proceedings of the 2021 Conference of the North American Chapter of the Association for Computational Linguistics: Human Language Technologies},
  pages     = {4599--4610},
  year      = {2021},
  publisher = {Association for Computational Linguistics},
  doi       = {10.18653/v1/2021.naacl-main.365}
}

@inproceedings{gu2024minillm,
  title     = {{MiniLLM}: Knowledge Distillation of Large Language Models},
  author    = {Gu, Yuxian and Dong, Li and Wei, Furu and Huang, Minlie},
  booktitle = {International Conference on Learning Representations},
  year      = {2024},
  url       = {https://openreview.net/forum?id=5h0qf7IBZZ}
}

@inproceedings{hsieh2023distilling,
  title     = {Distilling Step-by-Step! Outperforming Larger Language Models with Less Training Data and Smaller Model Sizes},
  author    = {Hsieh, Cheng-Yu and Li, Chun-Liang and Yeh, Chih-Kuan and Nakhost, Hootan and Fujii, Yasuhisa and Ratner, Alex and Krishna, Ranjay and Lee, Chen-Yu and Pfister, Tomas},
  booktitle = {Findings of the Association for Computational Linguistics: ACL 2023},
  pages     = {8003--8017},
  year      = {2023},
  publisher = {Association for Computational Linguistics},
  doi       = {10.18653/v1/2023.findings-acl.507}
}

@article{lanham2023measuring,
  title   = {Measuring Faithfulness in Chain-of-Thought Reasoning},
  author  = {Lanham, Tamera and Chen, Anna and Radhakrishnan, Ansh and Steiner, Benoit and Denison, Carson and Hernandez, Danny and Li, Dustin and Durmus, Esin and Hubinger, Evan and Kernion, Jackson and Lukosiute, Kamile and Nguyen, Karina and Cheng, Newton and Joseph, Nicholas and Schiefer, Nicholas and Rausch, Oliver and Larson, Robin and McCandlish, Sam and Kundu, Sandipan and Kadavath, Saurav and Yang, Shannon and Henighan, Thomas and Maxwell, Timothy and Telleen-Lawton, Timothy and Hume, Tristan and Hatfield-Dodds, Zac and Kaplan, Jared and Brauner, Jan and Bowman, Samuel R. and Perez, Ethan},
  journal = {arXiv preprint arXiv:2307.13702},
  year    = {2023},
  url     = {https://arxiv.org/abs/2307.13702}
}

@inproceedings{li2026demystifying,
  title     = {Demystifying Scientific Problem-Solving in {LLMs} by Probing Knowledge and Reasoning},
  author    = {Li, Alan and Liu, Yixin and Sarkar, Arpan and Downey, Doug and Cohan, Arman},
  booktitle = {International Conference on Machine Learning},
  year      = {2026},
  url       = {https://arxiv.org/abs/2508.19202}
}

@article{li2026rethinking,
  title   = {Rethinking On-Policy Distillation of Large Language Models: Phenomenology, Mechanism, and Recipe},
  author  = {Li, Yaxuan and Zuo, Yuxin and He, Bingxiang and Zhang, Jinqian and Xiao, Chaojun and Qian, Cheng and Yu, Tianyu and Gao, Huan-ang and Yang, Wenkai and Liu, Zhiyuan and Ding, Ning},
  journal = {arXiv preprint arXiv:2604.13016},
  year    = {2026}
}

@inproceedings{longpre2021entity,
  title     = {Entity-Based Knowledge Conflicts in Question Answering},
  author    = {Longpre, Shayne and Perisetla, Kartik and Chen, Anthony and Ramesh, Nikhil and DuBois, Chris and Singh, Sameer},
  booktitle = {Proceedings of the 2021 Conference on Empirical Methods in Natural Language Processing},
  pages     = {7052--7063},
  year      = {2021},
  publisher = {Association for Computational Linguistics},
  doi       = {10.18653/v1/2021.emnlp-main.565}
}

@inproceedings{ming2025faitheval,
  title     = {{FaithEval}: Can Your Language Model Stay Faithful to Context, Even If ``The Moon is Made of Marshmallows''?},
  author    = {Ming, Yifei and Purushwalkam, Senthil and Pandit, Shrey and Ke, Zixuan and Nguyen, Xuan-Phi and Xiong, Caiming and Joty, Shafiq},
  booktitle = {International Conference on Learning Representations},
  year      = {2025},
  url       = {https://openreview.net/forum?id=UeVx6L59fg}
}

@inproceedings{neeman2023disentqa,
  title     = {{DisentQA}: Disentangling Parametric and Contextual Knowledge with Counterfactual Question Answering},
  author    = {Neeman, Ella and Aharoni, Roee and Honovich, Or and Choshen, Leshem and Szpektor, Idan and Abend, Omri},
  booktitle = {Proceedings of the 61st Annual Meeting of the Association for Computational Linguistics (Volume 1: Long Papers)},
  pages     = {10056--10070},
  year      = {2023},
  publisher = {Association for Computational Linguistics},
  doi       = {10.18653/v1/2023.acl-long.559}
}

@inproceedings{turpin2023unfaithful,
  title     = {Language Models Don't Always Say What They Think: Unfaithful Explanations in Chain-of-Thought Prompting},
  author    = {Turpin, Miles and Michael, Julian and Perez, Ethan and Bowman, Samuel R.},
  booktitle = {Advances in Neural Information Processing Systems},
  volume    = {36},
  year      = {2023}
}

@article{yang2025qwen3,
  title   = {{Qwen3} Technical Report},
  author  = {Yang, An and Li, Anfeng and Yang, Baosong and Zhang, Beichen and Hui, Binyuan and Zheng, Bo and Yu, Bowen and Gao, Chang and Huang, Chengen and Lv, Chenxu and others},
  journal = {arXiv preprint arXiv:2505.09388},
  year    = {2025}
}

@inproceedings{ho2023reasoningteachers,
  title     = {Large Language Models Are Reasoning Teachers},
  author    = {Ho, Namgyu and Schmid, Laura and Yun, Se-Young},
  booktitle = {Proceedings of the 61st Annual Meeting of the Association for Computational Linguistics (Volume 1: Long Papers)},
  pages     = {14852--14882},
  year      = {2023},
  publisher = {Association for Computational Linguistics},
  doi       = {10.18653/v1/2023.acl-long.830}
}

@inproceedings{magister2023teaching,
  title     = {Teaching Small Language Models to Reason},
  author    = {Magister, Lucie Charlotte and Mallinson, Jonathan and Adamek, Jakub and Malmi, Eric and Severyn, Aliaksei},
  booktitle = {Proceedings of the 61st Annual Meeting of the Association for Computational Linguistics (Volume 2: Short Papers)},
  pages     = {1773--1781},
  year      = {2023},
  publisher = {Association for Computational Linguistics},
  doi       = {10.18653/v1/2023.acl-short.151}
}

@inproceedings{wang2023scott,
  title     = {{SCOTT}: Self-Consistent Chain-of-Thought Distillation},
  author    = {Wang, Peifeng and Wang, Zhengyang and Li, Zheng and Gao, Yifan and Yin, Bing and Ren, Xiang},
  booktitle = {Proceedings of the 61st Annual Meeting of the Association for Computational Linguistics (Volume 1: Long Papers)},
  pages     = {5546--5558},
  year      = {2023},
  publisher = {Association for Computational Linguistics},
  doi       = {10.18653/v1/2023.acl-long.304}
}

@inproceedings{deyoung2020eraser,
  title     = {{ERASER}: A Benchmark to Evaluate Rationalized {NLP} Models},
  author    = {DeYoung, Jay and Jain, Sarthak and Rajani, Nazneen Fatema and Lehman, Eric and Xiong, Caiming and Socher, Richard and Wallace, Byron C.},
  booktitle = {Proceedings of the 58th Annual Meeting of the Association for Computational Linguistics},
  pages     = {4443--4458},
  year      = {2020},
  publisher = {Association for Computational Linguistics},
  doi       = {10.18653/v1/2020.acl-main.408}
}

@inproceedings{ye2022unreliable,
  title     = {The Unreliability of Explanations in Few-shot Prompting for Textual Reasoning},
  author    = {Ye, Xi and Durrett, Greg},
  booktitle = {Advances in Neural Information Processing Systems},
  volume    = {35},
  pages     = {30378--30392},
  year      = {2022}
}

@article{rashkin2023attribution,
  title   = {Measuring Attribution in Natural Language Generation Models},
  author  = {Rashkin, Hannah and Nikolaev, Vitaly and Lamm, Matthew and Aroyo, Lora and Collins, Michael and Das, Dipanjan and Petrov, Slav and Tomar, Gaurav Singh and Turc, Iulia and Reitter, David},
  journal = {Computational Linguistics},
  volume  = {49},
  number  = {4},
  pages   = {777--840},
  year    = {2023},
  doi     = {10.1162/coli_a_00486}
}

@article{menick2022verifiedquotes,
  title   = {Teaching Language Models to Support Answers with Verified Quotes},
  author  = {Menick, Jacob and Trebacz, Maja and Mikulik, Vladimir and Aslanides, John and Song, Francis and Chadwick, Martin and Glaese, Mia and Young, Susannah and Campbell-Gillingham, Lucy and Irving, Geoffrey and McAleese, Nat},
  journal = {arXiv preprint arXiv:2203.11147},
  year    = {2022}
}

@inproceedings{gao2023citations,
  title     = {Enabling Large Language Models to Generate Text with Citations},
  author    = {Gao, Tianyu and Yen, Howard and Yu, Jiatong and Chen, Danqi},
  booktitle = {Proceedings of the 2023 Conference on Empirical Methods in Natural Language Processing},
  pages     = {6465--6488},
  year      = {2023},
  publisher = {Association for Computational Linguistics},
  doi       = {10.18653/v1/2023.emnlp-main.398}
}

@inproceedings{yue2023attribution,
  title     = {Automatic Evaluation of Attribution by Large Language Models},
  author    = {Yue, Xiang and Wang, Boshi and Chen, Ziru and Zhang, Kai and Su, Yu and Sun, Huan},
  booktitle = {Findings of the Association for Computational Linguistics: EMNLP 2023},
  pages     = {4615--4635},
  year      = {2023},
  publisher = {Association for Computational Linguistics},
  doi       = {10.18653/v1/2023.findings-emnlp.307}
}

@inproceedings{lewis2020rag,
  title     = {Retrieval-Augmented Generation for Knowledge-Intensive {NLP} Tasks},
  author    = {Lewis, Patrick and Perez, Ethan and Piktus, Aleksandra and Petroni, Fabio and Karpukhin, Vladimir and Goyal, Naman and K{\"u}ttler, Heinrich and Lewis, Mike and Yih, Wen-tau and Rockt{\"a}schel, Tim and Riedel, Sebastian and Kiela, Douwe},
  booktitle = {Advances in Neural Information Processing Systems},
  volume    = {33},
  pages     = {9459--9474},
  year      = {2020}
}

@inproceedings{asai2024selfrag,
  title     = {{Self-RAG}: Learning to Retrieve, Generate, and Critique through Self-Reflection},
  author    = {Asai, Akari and Wu, Zeqiu and Wang, Yizhong and Sil, Avirup and Hajishirzi, Hannaneh},
  booktitle = {International Conference on Learning Representations},
  year      = {2024},
  url       = {https://openreview.net/forum?id=hSyW5go0v8}
}

@inproceedings{kaushik2020counterfactual,
  title     = {Learning the Difference that Makes a Difference with Counterfactually-Augmented Data},
  author    = {Kaushik, Divyansh and Hovy, Eduard and Lipton, Zachary C.},
  booktitle = {International Conference on Learning Representations},
  year      = {2020},
  url       = {https://openreview.net/forum?id=Sklgs0NFvr}
}

@inproceedings{gardner2020contrastsets,
  title     = {Evaluating Models' Local Decision Boundaries via Contrast Sets},
  author    = {Gardner, Matt and Artzi, Yoav and Basmov, Victoria and Berant, Jonathan and Bogin, Ben and Chen, Sihao and Dasigi, Pradeep and Dua, Dheeru and Elazar, Yanai and Gottumukkala, Ananth and Gupta, Nitish and Hajishirzi, Hannaneh and Ilharco, Gabriel and Khashabi, Daniel and Lin, Kevin and Liu, Jiangming and Liu, Nelson F. and Mulcaire, Phoebe and Ning, Qiang and Singh, Sameer and Smith, Noah A. and Subramanian, Sanjay and Tsarfaty, Reut and Wallace, Eric and Zhang, Ally and Zhou, Ben},
  booktitle = {Findings of the Association for Computational Linguistics: EMNLP 2020},
  pages     = {1307--1323},
  year      = {2020},
  publisher = {Association for Computational Linguistics},
  doi       = {10.18653/v1/2020.findings-emnlp.117}
}

@inproceedings{lopezpaz2016privileged,
  title     = {Unifying Distillation and Privileged Information},
  author    = {Lopez-Paz, David and Bottou, L{\'e}on and Sch{\"o}lkopf, Bernhard and Vapnik, Vladimir},
  booktitle = {International Conference on Learning Representations},
  year      = {2016},
  url       = {https://arxiv.org/abs/1511.03643}
}

@inproceedings{wiegreffe2021association,
  title     = {Measuring Association Between Labels and Free-Text Rationales},
  author    = {Wiegreffe, Sarah and Marasovi{\'c}, Ana and Smith, Noah A.},
  booktitle = {Proceedings of the 2021 Conference on Empirical Methods in Natural Language Processing},
  year      = {2021},
  pages     = {10266--10284},
  publisher = {Association for Computational Linguistics},
  doi       = {10.18653/v1/2021.emnlp-main.804},
  url       = {https://aclanthology.org/2021.emnlp-main.804/}
}

@inproceedings{atanasova2023faithfulness,
  title     = {Faithfulness Tests for Natural Language Explanations},
  author    = {Atanasova, Pepa and Camburu, Oana-Maria and Lioma, Christina and Lukasiewicz, Thomas and Simonsen, Jakob Grue and Augenstein, Isabelle},
  booktitle = {Proceedings of the 61st Annual Meeting of the Association for Computational Linguistics (Volume 2: Short Papers)},
  year      = {2023},
  pages     = {283--294},
  publisher = {Association for Computational Linguistics},
  doi       = {10.18653/v1/2023.acl-short.25},
  url       = {https://aclanthology.org/2023.acl-short.25/}
}

@article{chen2025reasoning,
  title   = {Reasoning Models Don't Always Say What They Think},
  author  = {Chen, Yanda and Benton, Joe and Radhakrishnan, Ansh and Uesato, Jonathan and Denison, Carson and Schulman, John and Somani, Arushi and Hase, Peter and Wagner, Misha and Roger, Fabien and Mikulik, Vlad and Bowman, Samuel R. and Leike, Jan and Kaplan, Jared and Perez, Ethan},
  journal = {arXiv preprint arXiv:2505.05410},
  year    = {2025},
  url     = {https://arxiv.org/abs/2505.05410}
}

\clearpage
\appendix
\section{Training and implementation details}
\label{app:training}

\subsection{Optimization}

\begin{table}[!t]
\caption{Training and decoding configuration. Pair hyperparameters are selected once on \qasper development data and transferred unchanged.}
\label{tab:hyperparameters}
\centering
\small
\begin{tabular}{L{0.46\linewidth}L{0.42\linewidth}}
\toprule
\textbf{Setting} & \textbf{Value} \\
\midrule
Student checkpoints & Qwen3-1.7B-Base; Qwen3-4B-Base \\
Frozen teacher & Qwen3-30B-A3B-Instruct-2507 \\
Trainable parameters & all student parameters \\
Optimizer & AdamW, $\beta=(0.9,0.95)$, weight decay $0.1$ \\
Peak learning rate & $1.5\times10^{-5}$ (1.7B); $1.0\times10^{-5}$ (4B) \\
Schedule & 3\% warmup, cosine decay, two epochs \\
Global batch & 128 documents (64 synchronized pairs for \method) \\
Maximum input / output & 32,768 / 768 tokens \\
Training rollout & temperature $0.8$, top-$p=0.95$ \\
Primary evaluation & greedy decoding \\
Gradient clipping / precision & $1.0$ / bfloat16 \\
OPD divergence & token-average reverse KL, full vocabulary \\
Pair margin $m$ / weight $\lambda$ & $0.20$ / $0.50$ \\
Random seeds & 3 per principal configuration \\
\bottomrule
\end{tabular}
\end{table}

All methods use the same document serialization and answer delimiter. The frozen evidence-preserving pack keeps the retrieval channel fixed across methods, separating evidence use from retrieval-augmented generation~\citep{lewis2020rag,asai2024selfrag}. OPD and \method compute the teacher distribution at every non-padding student token; no reward model or answer-only policy gradient is used. To match supervision tokens, we subsample longer off-policy traces and resample shorter answer-only targets. Training examples are shuffled at the paper level. Development selection uses original accuracy subject to a counterfactual-response constraint fixed before test evaluation; the test checkpoint is selected once per seed.

For \method, a batch contains the original and counterfactual member of each pair. The global batch size counts documents, so the number of pairs per update is half the number of standard-OPD documents. Both members receive independent on-policy rollouts. The margin score is teacher-forced only over the short canonical answer string and is length normalized as in \cref{eq:pairloss}.

\subsection{Evidence-Paired OPD details}
\label{app:epopd-details}

Let $x=(q,D)$ and let the student sample $y\sim\student(\cdot\mid x)$. At each student-visited prefix $y_{<t}$, the frozen teacher provides a next-token distribution $\teacher(\cdot\mid x,y_{<t})$. The standard baseline uses length-normalized reverse KL,
\begin{equation}
\loss_{\mathrm{OPD}}(x;\theta)=
\E_{y\sim\student(\cdot\mid x)}\!\left[
\frac{1}{|y|}\sum_{t=1}^{|y|}
\KL\!\left(\student(\cdot\mid x,y_{<t})\,\|\,
\teacher(\cdot\mid x,y_{<t})\right)\right].
\label{eq:opd-app}
\end{equation}
For a validated pair $x^0=(q,D^0)$ and $x^1=(q,D^1)$, the answer score used in the margin is
\begin{equation}
s_\theta(a\mid x)=\frac{1}{|a|}\sum_{t=1}^{|a|}
\log \student(a_t\mid x,a_{<t}),
\end{equation}
and the complete training objective is
\begin{equation}
\loss_{\mathrm{EP\text{-}OPD}}=
\tfrac12\sum_{b\in\{0,1\}}\loss_{\mathrm{OPD}}(x^b;\theta)
+\lambda\loss_{\mathrm{pair}},
\end{equation}
with $\loss_{\mathrm{pair}}$ defined in \cref{eq:pairloss}.

The complete update is summarized in \cref{alg:epopd} in the main text.

\subsection{Compute matching}
\label{c3}
Teacher forward calls dominate OPD compute. Standard OPD therefore receives two independently sampled observational documents for each pair consumed by \method, equalizing teacher-scored tokens and optimizer steps. Reported method comparisons also match maximum input length, maximum generated length, decoding temperature, and checkpoint-selection schedule. Where EP-OPD is evaluated at both student sizes, the runs share data manifests and hyperparameters except for learning rate.

\subsection{Factorized OPD control objectives}
\label{app:factor-controls}

Every factor control starts from the same \qasper--4B training manifest and changes only the specified training choice. EMA targets are no-gradient moving teachers; if $\phi_k$ denotes the corresponding online teacher checkpoint after update $k$, their parameters obey
\begin{equation}
\bar\phi_k=\mu\bar\phi_{k-1}+(1-\mu)\phi_k,
\qquad \mu\in\{0.995,0.95\},
\end{equation}
with identical initialization and update ordering within the two EMA runs. The fixed-teacher row instead uses the external Qwen3-30B-A3B target throughout.

For privileged-context controls, the student's context remains $x=(q,D)$. The teacher receives $x_T=(q,D,z_E)$, where weak PI makes $z_E$ the IDs of gold evidence paragraphs and strong PI makes $z_E$ the exact answer-critical evidence span. PI is never present in the student rollout or at evaluation. Because these rows consume information unavailable to the symmetric-context methods, they are oracle interventions in the learning-with-privileged-information sense~\citep{lopezpaz2016privileged} and estimate the value of resolving the teacher's evidence-location uncertainty.

At a student-visited prefix $s_t=(x,y_{<t})$, forward KL uses
\begin{equation}
\loss_{\mathrm{FKL}}(s_t;T)=T^2\KL\!\left(
\teacher^{(T)}(\cdot\mid s_t)\,\|\,\student^{(T)}(\cdot\mid s_t)\right),
\end{equation}
where both distributions use temperature $T$ and the conventional $T^2$ factor compensates for temperature-dependent gradient scaling. Entropy masks are computed from the frozen pre-update student distribution for the complete rollout. Let $m_t^{(q)}$ select prefixes in the top $q\in\{0.50,0.25\}$ entropy quantile. The masked loss is
\begin{equation}
\loss_{\mathrm{mask}}^{(q)}=
\frac{\sum_t m_t^{(q)}\loss_{\mathrm{OPD}}(s_t)}
{\sum_t m_t^{(q)}+\epsilon}.
\end{equation}
Normalizing over the selected prefixes keeps the loss an average over supervised positions. The ablation therefore changes which positions receive supervision, rather than reducing the loss simply because fewer positions are selected. Rollouts and optimizer steps are held fixed; teacher-scored tokens intentionally fall with $q$ because supervision density is the manipulated variable.

\section{Comparison and supporting results}
\label{app:proofs}

\subsection{Metric definitions and paired training comparison}
\label{app:formal-comparison}

Let $C_i^v(\theta)=\ind[\hat a_\theta(q_i,D_i^v)=a_i^v]$ denote answer correctness on document version $v\in\{\mathrm{orig},\mathrm{cf}\}$. The primary answer-level metrics are
\begin{align}
 \Acc_{\mathrm{orig}}(\theta) &= \E_i[C_i^{\mathrm{orig}}(\theta)],\\
 \CFAcc(\theta) &= \E_i[C_i^{\mathrm{cf}}(\theta)],\\
 \Bypass(\theta) &=
 \Pr\!\left[\hat a_\theta(q_i,D_i^{\mathrm{cf}})=a_i^{\mathrm{orig}}
 \mid C_i^{\mathrm{orig}}(\theta)=1\right].
 \label{eq:metrics-app}
\end{align}
Original accuracy measures performance on the observed document, CFAcc measures adaptation to the answer-changing document, and bypass measures persistence of the original answer among examples answered correctly before the intervention.

For a training method $M$, with base parameters $\theta_0$ and trained parameters $\theta_M$, define
\begin{align}
 \Delta_M^v &= \E_i\!\left[C_i^v(\theta_M)-C_i^v(\theta_0)\right],\\
 \Psi_M &= \Delta_M^{\mathrm{cf}}-\Delta_M^{\mathrm{orig}}.
 \label{eq:estimand}
\end{align}
The paired quantity $\Psi_M$ measures how the training-induced change differs between counterfactual and original documents. We report it together with the corresponding change in bypass to summarize how training changes document response.


\subsection{Single-document ambiguity}

\begin{proposition}[Observational non-identifiability]
\label[proposition]{prop:nonident}
Let $\mathcal S=\{(q_i,D_i,a_i)\}_{i=1}^n$ be a training support in which each $q_i$ occurs with one answer-supporting document. If an evidence-responsive policy $f(q,D)$ is correct on $\mathcal S$, then there exists a question-only policy $g(q)$ with identical predictions and identical empirical answer or trace-imitation risk on $\mathcal S$. Consequently, observational supervision alone does not identify evidence response.
\end{proposition}

\begin{proof}
Let $f$ be any evidence-responsive policy that produces a conditional output distribution $p_f(\cdot\mid q_i,D_i)$ on each observed example. Because each observed $q_i$ is paired with a single $D_i$, define a question-only policy $g$ on the empirical support by
\begin{equation}
p_g(\cdot\mid q_i)=p_f(\cdot\mid q_i,D_i).
\end{equation}
Outside the observed questions, define $g$ arbitrarily. Then $f$ and $g$ induce identical distributions over answers on $\mathcal S$, hence identical empirical answer loss for every proper scoring rule. If their next-token distributions are matched at every prefix reached on $\mathcal S$, they also induce the same trace-imitation and on-policy token risk there. Nevertheless, $g$ is invariant to replacing $D_i$ with any $D_i'$, whereas $f$ need not be. The two policies therefore agree observationally but differ under document intervention.
\end{proof}

The proposition characterizes the observed training support: one document per question leaves evidence-responsive and question-conditioned solutions observationally indistinguishable on that support.

\begin{proposition}[Paired shortcut exclusion]
\label[proposition]{prop:pairlower}
Consider a balanced pair with fixed $q$, distinct answers $a^0\neq a^1$, and documents $D^0,D^1$. Any stochastic question-only policy $p(a\mid q)$ has average negative log-likelihood at least $\log 2$ on the pair. Any deterministic question-only policy has pair accuracy at most $1/2$.
\end{proposition}

\begin{proof}
Write $p_0=p(a^0\mid q)$ and $p_1=p(a^1\mid q)$. Since $a^0\neq a^1$, $p_0+p_1\leq1$. The balanced pair negative log-likelihood is
\begin{equation}
\ell=-\tfrac12(\log p_0+\log p_1)=-\log\sqrt{p_0p_1}.
\end{equation}
By AM--GM, $\sqrt{p_0p_1}\leq(p_0+p_1)/2\leq1/2$, so $\ell\geq\log2$. A deterministic question-only policy returns the same answer for both documents, so it can match at most one of the two distinct labels.
\end{proof}

For a balanced $K$-way evidence set with $K$ distinct supported answers, the same argument gives NLL at least $\log K$ and deterministic accuracy at most $1/K$.

\subsection{What the paired margin guarantees locally}
\label{a3}
\begin{corollary}[Local evidence-response certificate]
\label[corollary]{cor:certificate}
If $\loss_{\mathrm{pair}}=0$, then the paired response
\begin{equation}
R_\theta=\big[\score(a^0\mid x^0)-\score(a^1\mid x^0)\big]
+\big[\score(a^1\mid x^1)-\score(a^0\mid x^1)\big]
\end{equation}
satisfies $R_\theta\geq2m$. Every question-only policy has $R_\theta=0$, so it cannot attain zero pair loss for $m>0$.
\end{corollary}

\begin{proof}
Zero hinge loss in \cref{eq:pairloss} implies
\begin{align}
\score(a^0\mid x^0)-\score(a^1\mid x^0)&\geq m,\\
\score(a^1\mid x^1)-\score(a^0\mid x^1)&\geq m.
\end{align}
Summing yields $R_\theta\geq2m$. For a question-only policy, $\score(a\mid q,D^0)=\score(a\mid q,D^1)$ for every $a$, making the two brackets exact negatives and $R_\theta=0$.
\end{proof}

\begin{corollary}[Paired answer-decision guarantee]
\label[corollary]{cor:decision}
Let $d(x)=\arg\max_{a\in\{a^0,a^1\}}\score(a\mid x)$. If $\loss_{\mathrm{pair}}=0$ and $m>0$, then $d(x^0)=a^0$ and $d(x^1)=a^1$. Consequently, candidate-set accuracy is one on both documents and pairwise bypass is zero.
\end{corollary}

\begin{proof}
Zero pair loss with $m>0$ implies $\score(a^0\mid x^0)>\score(a^1\mid x^0)$ and $\score(a^1\mid x^1)>\score(a^0\mid x^1)$. The candidate-set argmax therefore selects the document-supported answer on both versions.
\end{proof}

These statements characterize the two canonical alternatives used by the margin. The main evaluation extends the analysis to fresh document pairs under free generation.

\subsection{A simple illustration of prior cancellation}

\begin{proposition}[Prior cancellation in a two-source model]
\label[proposition]{prop:cancellation}
Let $c\in\{-1,+1\}$ index which of two answers a document supports, and let the logit for the positive answer be $u(q)+cv$, where $u(q)$ is constant across the pair and $v>0$ is the evidence contribution. For fixed $v$, the balanced paired logistic risk has the unique minimizer $u^*=0$. Observational risk on only $c=+1$ depends on $u+v$ and therefore cannot identify their allocation.
\end{proposition}

\begin{proof}
For the fixed question, write $u=u(q)$. The balanced paired logistic risk is
\begin{equation}
L(u,v)=\tfrac12\log(1+e^{-(u+v)})+\tfrac12\log(1+e^{u-v}).
\end{equation}
For fixed $v$, $L$ is strictly convex in $u$, symmetric under $u\mapsto-u$, and has $\partial L/\partial u=0$ at $u=0$; hence its unique minimizer is $u^*=0$. With only the positive document, the risk is $\log(1+e^{-(u+v)})$, which is constant on every level set of $u+v$ and cannot identify how the logit is allocated.
\end{proof}

Writing $\sigma(z)=(1+e^{-z})^{-1}$, original-answer persistence under the counterfactual document is $B(u,v)=\sigma(u-v)$. At fixed $v$,
\begin{align}
\frac{\partial B}{\partial u}
&=\sigma(u-v)\bigl[1-\sigma(u-v)\bigr]>0,\\
\frac{\partial L}{\partial u}
&=\tfrac12\bigl[\sigma(u-v)-\sigma(-u-v)\bigr]>0
\qquad (u>0),\\
\frac{\partial^2 L}{\partial u^2}
&=\tfrac12\bigl[\sigma'(u-v)+\sigma'(-u-v)\bigr]>0.
\end{align}
Thus a larger positive prior increases persistence and also strengthens the paired gradient opposing it in this toy model. This relation motivates the pre-training prior stratification used in the main experiments.

\section{Evidence intervention construction}
\label{app:construction}

\subsection{Construction pipeline}

Our minimal, label-changing revisions follow the local-control principle of counterfactually augmented data and contrast sets~\citep{kaushik2020counterfactual,gardner2020contrastsets}. For every answerable question with annotated evidence, we apply the following deterministic-first pipeline.

\begin{enumerate}
    \item \textbf{Evidence closure.} Start from the annotated evidence spans and add coreferent or explicitly dependent sentences needed to interpret the target claim. This prevents an edit from leaving a contradictory duplicate elsewhere in the local context.
    \item \textbf{Atomic answer graph.} Normalize the gold answer into typed slots (entity, scalar with unit, signed direction, Boolean polarity, comparison, or bounded set). Link each slot to the smallest evidence statement that determines it.
    \item \textbf{Type-preserving proposal.} Apply an operator from \cref{tab:operators}. The proposed value is selected to remain locally plausible, preserve grammar and units, and differ from every answer alias in the original document.
    \item \textbf{Dependent rewrite.} Propagate the edit only through the evidence closure. Section headings, citations, surrounding methods, distractor paragraphs, and the question are frozen.
    \item \textbf{Automatic validation.} Require answer change, type validity, lexical-diff locality, no remaining original-support span, no newly introduced contradiction, and agreement between the deterministic answer transformer and a document-only verifier.
    \item \textbf{Held-out validation.} Review the complete evaluation pair without revealing training method or model outputs. Reject rather than repair ambiguous pairs after model evaluation begins.
\end{enumerate}

\begin{table}[!t]
\caption{Counterfactual edit operators. All edits are restricted to the evidence closure and preserve answer type.}
\label{tab:operators}
\centering
\small
\begin{tabular}{L{0.18\linewidth}L{0.31\linewidth}L{0.41\linewidth}}
\toprule
\textbf{Operator} & \textbf{Example transformation} & \textbf{Required validation} \\
\midrule
Entity & Method A $\rightarrow$ Method B & same entity class; no residual coreference \\
Numeric & $4.2\rightarrow2.4$ points & unit and precision preserved; table/text agree \\
Direction & improves $\rightarrow$ degrades & predicate antonym valid in local syntax \\
Polarity & supported $\rightarrow$ not supported & all scoped negation dependencies updated \\
Comparison & A $>$ B $\rightarrow$ B $>$ A & metric and comparison set unchanged \\
Bounded set & \{A,B\} $\rightarrow$ \{A,C\} & cardinality and entity types preserved \\
\bottomrule
\end{tabular}
\end{table}

\subsection{Control documents}

The four modified document versions distinguish sensitivity to answer-critical evidence from sensitivity to changes that should leave the answer unchanged.

\begin{itemize}
    \item $\Dpara$: paraphrase only the answer-critical evidence while preserving every typed answer slot. This tests semantic invariance.
    \item $\Dirr$: insert paragraphs from another paper matched in length, section type, and broad topic but unrelated to $q$. This tests distraction and context length.
    \item $\Ddrop$: remove the full evidence closure and repair only dangling discourse markers. The correct behavior is to state that the document is insufficient.
    \item $\Dcf$: edit the answer-critical slots and all dependent mentions so that the document uniquely supports $a^{\mathrm{cf}}$.
\end{itemize}

The context tokenizer length is matched between $\Dorig$ and $\Dcf$ up to the small unavoidable difference introduced by type-preserving replacements. For the conditional bypass computation, answer aliases are canonicalized before comparing a counterfactual output with $a^{\mathrm{orig}}$.

\section{Evaluation protocol}
\label{app:evaluation}

\subsection{Deterministic answer scoring}

Each primary item stores its answer type, canonical value, and allowed aliases. Text is Unicode-normalized, case-folded where appropriate, and stripped of citation markers before comparison. Numeric values are parsed with units and converted to the annotated unit; no tolerance is allowed unless the source itself reports rounding. Boolean, direction, and comparison answers use a fixed ontology. Set-valued answers require exact normalized set equality. Outputs that provide both the original and counterfactual answer without resolving the question are incorrect. This restricted atomic scorer is why headline $\Acc$ and $\CFAcc$ do not require an LLM judge.

\subsection{Atomic claim segmentation}

Claim-level evaluation asks whether a response is supported by its cited source, as in attribution and citation evaluation~\citep{rashkin2023attribution,menick2022verifiedquotes,gao2023citations,yue2023attribution}. Although the evaluation prompt requests one claim per step, we segment the output before assigning support labels. Coordinated propositions are split when either can be contradicted independently, and cited premises are separated from a derived conclusion. An equation, its substituted values, and its verbal interpretation are treated as separate units when each can fail independently; discourse markers and restatements are omitted.

Segmentation uses anonymized outputs without method, checkpoint, document-version, or answer-correctness metadata. Boundary cases are resolved with a fixed guide before support labels are revealed. This procedure reduces the extent to which differences in verbosity or sentence structure change the unit being evaluated. These labels assess source support, not whether the claims describe the model's internal reasoning.

\subsection{Structured evidence response prompt}

\begin{center}
\fcolorbox{black!55}{SoftGray}{%
\begin{minipage}{0.91\linewidth}
\raggedright
\ttfamily\small
You are answering a question using only the supplied scientific document.\\
For each reasoning step, return exactly:\\
CLAIM: <one atomic claim>\\
EVIDENCE: <paragraph IDs that support this claim, or NONE>\\
RELATION: direct | composed | uncertain\\
After the steps, return:\\
ANSWER: <answer, or INSUFFICIENT EVIDENCE>\\[2pt]
Do not use facts that are absent from the document. If the necessary evidence is missing, abstain.
\end{minipage}}
\end{center}

All systems receive this response contract only at evaluation; it is not a method-specific prompt. Paragraph identifiers are assigned before document variants are generated so that unchanged paragraphs retain the same IDs.

\section{Paired-document component ablation}
\label{app:paired_ablation}

\paragraph{Configurations.}
We examine the contributions of counterfactual document exposure,
answer supervision, and the paired margin on \qasper with the
Qwen3-4B student. Each paired-document configuration uses original
and counterfactual documents supporting different answers to the
same question. Let
\[
\mathcal{L}_{\mathrm{dist}}
=
\frac{1}{2}\sum_{b\in\{0,1\}}
\mathcal{L}_{\mathrm{OPD}}(x^b;\theta)
\]
denote the paired-document distillation term, and let
$\mathcal{L}_{\mathrm{CE}}$ denote cross-entropy supervision of each
document's supported canonical answer.
The compared objectives are
\[
\begin{aligned}
\mathcal{L}_{\mathrm{Augmented\ OPD}}
    &= \mathcal{L}_{\mathrm{dist}},\\
\mathcal{L}_{\mathrm{Paired\ ANS\text{-}SFT}}
    &= \mathcal{L}_{\mathrm{CE}},\\
\mathcal{L}_{\mathrm{Augmented\ OPD+CE}}
    &= \mathcal{L}_{\mathrm{dist}}
       + \gamma\mathcal{L}_{\mathrm{CE}},\\
\mathcal{L}_{\mathrm{EP\text{-}OPD}}
    &= \mathcal{L}_{\mathrm{dist}}
       + \lambda\mathcal{L}_{\mathrm{pair}},
\end{aligned}
\]
where $\gamma$ weights answer cross-entropy.
Augmented OPD removes the answer-level objective from EP-OPD;
Paired ANS-SFT uses answer supervision without distillation; and
Augmented OPD + Answer CE replaces the paired margin with
cross-entropy. Standard OPD provides the original-document reference.

\begin{table}[!t]
    \centering
    \small
    \setlength{\tabcolsep}{6pt}
    \caption{
        Paired-document component ablation on \qasper with Qwen3-4B.
        Scores are percentages. Standard OPD uses original documents;
        the remaining configurations use original and counterfactual
        documents. Bold denotes the best value in each column.
    }
    \label{tab:paired_ablation}
    \begin{tabular}{@{}lrrrr@{}}
        \toprule
        \textbf{Method}
        & \textbf{Orig.$\uparrow$}
        & \textbf{CFAcc$\uparrow$}
        & \textbf{Bypass$\downarrow$}
        & \textbf{CSP$\uparrow$} \\
        \midrule
        Standard OPD
        & 52.4 & 39.2 & 35.8 & 57.0 \\
        Augmented OPD
        & 52.1 & 38.5 & 27.2 & 62.3 \\
        Paired ANS-SFT
        & 50.5 & 42.0 & \textbf{25.3} & 63.8 \\
        Augmented OPD + Answer CE
        & \textbf{52.5} & 40.9 & 26.7 & 59.7 \\
        \midrule
        EP-OPD
        & 51.9 & \textbf{42.7} & 25.9 & \textbf{65.4} \\
        \bottomrule
    \end{tabular}
\end{table}

\paragraph{Document exposure and the paired margin.}
Table~\ref{tab:paired_ablation} shows that adding counterfactual
documents changes the evidence-response profile: Augmented OPD
reduces bypass from 35.8\% to 27.2\% and increases CSP from 57.0\%
to 62.3\%, while CFAcc changes from 39.2\% to 38.5\%.
Lower original-answer persistence thus accompanies a different
change in counterfactual correctness, highlighting the complementary
information provided by these metrics.
Adding the paired margin increases CFAcc from 38.5\% to 42.7\%,
raises CSP from 62.3\% to 65.4\%, and further reduces bypass to
25.9\%, with original accuracy changing from 52.1\% to 51.9\%.

\paragraph{Answer supervision and objective choice.}
Paired ANS-SFT achieves the lowest bypass rate, 25.3\%.
Relative to this configuration, EP-OPD raises original accuracy
by 1.4 percentage points, CFAcc by 0.7 points, and CSP by 1.6 points,
with a 0.6-point increase in bypass.
Combining augmented OPD with answer cross-entropy yields the
highest original accuracy, 52.5\%.
Compared with this objective, EP-OPD improves CFAcc by 1.8 points
and CSP by 5.7 points, while reducing bypass by 0.8 points at
a 0.6-point original-accuracy cost.
These comparisons characterize the joint contribution of
on-policy supervision and document-specific answer ranking:
the complete objective attains the highest CFAcc and CSP in
the ablation while retaining an original accuracy of 51.9\%.



\section{Additional results}
\label{app:results}

\subsection{Document-operator localization}

\begin{table}[!t]
\caption{Document-operator and claim-level outcomes on \qasper--4B. Changes compare standard OPD with the frozen base. Scores are percentages, except the paraphrase gap and the change column, which are in pp. The expert row uses the blinded validation subset.}
\label{tab:grounding-controls}
\centering
\small
\setlength{\tabcolsep}{5pt}
\begin{tabularx}{\linewidth}{@{}>{\raggedright\arraybackslash}Xrrr@{}}
\toprule
\textbf{Outcome} & \textbf{Base} & \textbf{Standard OPD} & \textbf{OPD $-$ Base} \\
\midrule
Original-document accuracy & \BaseOrig & \OpdOrig & $+\MainOrigGain$ \\
Paraphrase accuracy & \BasePara & \OpdPara & $+3.2$ \\
Paraphrase gap (Para. $-$ Orig.) & $-0.3$ & $-0.5$ & $-0.2$ \\
Counterfactual accuracy & \BaseCF & \OpdCF & $\MainCFLoss$ \\
Evidence-drop abstention & \BaseDrop & \OpdDrop & $-7.6$ \\
Conditional bypass & \BaseBypass & \OpdBypass & $+\MainBypassGain$ \\
\midrule
Automatic step evidence error & \JudgeErrBase & \JudgeErrOpd & $+\JudgeErrDelta$ \\
Expert step evidence error & \ExpertErrBase & \ExpertErrOpd & $+\ExpertErrDelta$ \\
\bottomrule
\end{tabularx}
\end{table}

Answer-preserving paraphrase and irrelevant-insertion controls retain most of the observed-document gain. Evidence deletion and answer-changing interventions produce the largest changes in evidence-sensitive outcomes, and the claim-level rows show the same source-support pattern.

\subsection{Parametric-prior strata}
\label{f2}

\begin{table}[!t]
\caption{Bypass rate (\%) by frozen-base parametric prior quartile. Parentheses show the change from base.}
\label{tab:prior-quartiles}
\centering
\begin{tabular}{lrrrr}
\toprule
\textbf{Method} & \textbf{Lowest} & \textbf{Q2} & \textbf{Q3} & \textbf{Highest} \\
\midrule
\base & 12 & 16 & 21 & 29 \\
Standard \opd & 17 ($+5$) & 24 ($+8$) & 34 ($+13$) & 49 ($+20$) \\
\method & 14 ($+2$) & 19 ($+3$) & 25 ($+4$) & 35 ($+6$) \\
\bottomrule
\end{tabular}
\end{table}

The base model already shows greater original-answer persistence in the stronger-prior quartiles. Our main comparison concerns the additional increase after OPD, which grows from $+5$ to $+20$ pp across the quartiles. \method reduces most of this training-induced increase, while the differences already present in the base model remain.

\subsection{Empirical Diagnostics}

\begin{table}[!t]
\caption{Implications and observed outcomes. Changes are relative to the frozen base unless a repair comparison is stated; main-setting results refer to \qasper--4B.}
\label{tab:criteria}
\centering
\small
\setlength{\tabcolsep}{4pt}
\renewcommand{\arraystretch}{1.15}
\begin{tabularx}{\linewidth}{@{}L{0.29\linewidth}L{0.32\linewidth}>{\raggedright\arraybackslash}X@{}}
\toprule
\textbf{Implication} & \textbf{Expected direction} & \textbf{Observed} \\
\midrule
1. Observational competence rises
& $\Delta_{\rm OPD}^{\rm orig}>0$
& $+\MainOrigGain$ pp (main); $+2.1$ to $+3.4$ pp across settings \\
2. Evidence response separates
& $\Psi_{\rm OPD}<0$
& $\MainInteraction$ pp (main); $\Delta\CFAcc=\MainCFLoss$ pp \\
3. Original answer persists
& $\Delta\Bypass>0$
& $+\MainBypassGain$ pp (main) \\
4. Reallocation replicates and increases with prior strength
& Same signs across settings; larger bypass increment at higher prior
& Three other settings; quartile increments $+5$ to $+20$ pp \\
5. Pairing restores evidence response
& $\CFAcc_{\rm EP}>\CFAcc_{\rm OPD}$
& $54$--$66\%$ of CFAcc loss recovered \\
6. Repair preserves the original-task benefit
& $\Acc_{\rm orig}^{\rm EP}>\Acc_{\rm orig}^{\rm Base}$
& $80$--$85\%$ of gain retained; $\EpOrigCost$ pp vs.\ OPD (main) \\
7. Claim support tracks reallocation and repair
& Evidence error rises after OPD; CSP rises with repair
& Expert error $+\ExpertErrDelta$ pp; repair CSP $+\EpCSPGain$ pp (main) \\
\bottomrule
\end{tabularx}
\end{table}

These comparisons bring together the main finding, its replication, and the effect of evidence pairing. The first four describe how answer accuracy and evidence response change after OPD, including their relationship to the student's initial answer preference. The remaining comparisons assess whether \method improves evidence response and claim support while retaining the original-task gain. Together with the document controls in \cref{tab:grounding-controls}, they help assess alternatives such as general perturbation sensitivity or indiscriminate answer suppression. The direction column summarizes the qualitative relationships under examination; the observed column gives the reported magnitudes.



\section{Illustrative intervention examples}
\label{app:examples}

\paragraph{Direction edit.}
``The proposed regularizer \emph{improves} calibration'' becomes ``\emph{worsens} calibration.'' The question and experiment description are unchanged; the correct comparison reverses.

\paragraph{Numeric edit.}
``Ablation reduces F1 by $4.2$ points'' becomes ``$0.4$ points.'' Units, table position, and qualitative framing are preserved.

\paragraph{Evidence deletion.}
Sentences locating an out-of-domain split are removed while dataset names remain in unrelated paragraphs. A supported system must abstain.

These schematic examples use synthetic text for illustration. The manifest records each exact diff, answer transformation, evidence closure, and validation status.

\end{document}